\documentclass[9pt, conference]{IEEEtran}
\IEEEoverridecommandlockouts

\usepackage{cite}
\usepackage{amsmath,graphicx}
\usepackage[hidelinks]{hyperref}
\usepackage{subfigure}
\usepackage{tikz}
\usepackage{pgfplots}
\usepackage{booktabs}
\usepackage{multirow}
\usepackage{array}
\usepackage{utfsym}
\usepackage{amssymb}
\usepackage{amsfonts}
\pgfplotsset{compat=1.18}
\usepackage{authblk}

\newcommand{\PreserveBackslash}[1]{\let\temp=\\#1\let\\=\temp}

\newcolumntype{C}[1]{>{\PreserveBackslash\centering}p{#1}}
\newcolumntype{R}[1]{>{\PreserveBackslash\raggedleft}p{#1}}
\newcolumntype{L}[1]{>{\PreserveBackslash\raggedright}p{#1}}

\usepackage[linesnumbered,ruled,vlined]{algorithm2e}

\SetCommentSty{mycommfont}

\SetKwInput{KwInput}{Input}                
\SetKwInput{KwOutput}{Output}              
\SetKwInput{KwInit}{Definition} 

\def\BibTeX{{\rm B\kern-.05em{\sc i\kern-.025em b}\kern-.08em
    T\kern-.1667em\lower.7ex\hbox{E}\kern-.125emX}}

\begin{document}

\title{A Modular-based Strategy for Mitigating Gradient Conflicts in Simultaneous Speech Translation
\thanks{
This work was supported in part by the National Science Foundation of China (Nos. 62276056 and U24A20334), the Natural Science Foundation of Liaoning Province of China (2022-KF-26-01), the Fundamental Research Funds for the Central Universities (Nos. N2216016 and N2316002), the Yunnan Fundamental Research Projects (No. 202401BC070021), and the Program of Introducing Talents of Discipline to Universities, Plan 111 (No.B16009).\\
$\ast$ Equal contribution.\\
$\dagger$ Corresponding author.}
}

\author{
  Xiaoqian Liu\textsuperscript{1$\ast$}, Yangfan Du\textsuperscript{1$\ast$}, Jianjin Wang\textsuperscript{1}, Yuan Ge\textsuperscript{1}, Chen Xu\textsuperscript{2}, Tong Xiao\textsuperscript{1,3$\dagger$}, Guocheng Chen\textsuperscript{1} and Jingbo Zhu\textsuperscript{1,3}\\
  \textsuperscript{1}School of Computer Science and Engineering, Northeastern University, Shenyang, China\\
  \textsuperscript{2}College of Computer Science and Technology, Harbin Engineering University, Harbin, China\\
  \textsuperscript{3}NiuTrans Research, Shenyang, China
}

\maketitle

\begin{abstract}

Simultaneous Speech Translation (SimulST) involves generating target language text while continuously processing streaming speech input, presenting significant real-time challenges. 
Multi-task learning is often employed to enhance SimulST performance but introduces optimization conflicts between primary and auxiliary tasks, potentially compromising overall efficiency. The existing model-level conflict resolution methods are not well-suited for this task which exacerbates inefficiencies and leads to high GPU memory consumption. To address these challenges, we propose a \textit{Modular Gradient Conflict Mitigation (MGCM)} strategy that detects conflicts at a finer-grained modular level and resolves them utilizing gradient projection. Experimental results demonstrate that MGCM significantly improves SimulST performance, particularly under medium and high latency conditions, achieving a 0.68 BLEU score gain in offline tasks. Additionally, MGCM reduces GPU memory consumption by over 95\% compared to other conflict mitigation methods, establishing it as a robust solution for SimulST tasks.

\end{abstract}

\begin{IEEEkeywords}
Simultaneous Speech Translation, Multi-task Learning, Gradient Conflict Mitigation, GPU Memory Efficiency
\end{IEEEkeywords}

\section{Introduction}

End-to-end Simultaneous Speech Translation (SimulST)\cite{fugen2007simultaneous, oda2014optimizing, liu2024recent} is a critical technology that generates target language text while continuously receiving streaming speech input. Unlike traditional offline speech translation\cite{ney1999speech, xu2023recent, xu2021stacked}, SimulST requires real-time processing, necessitating not only cross-lingual and cross-modal modeling but also ensuring translation accuracy and fluency despite incomplete information. This complex modeling requirement often leads to insufficient learning and suboptimal performance \cite{liu2024recent}.

One of the most common approaches to enhance SimulST model performance is Multi-task Learning (MTL)\cite{xu2023ctc}, which integrates auxiliary tasks such as Simultaneous Automatic Speech Recognition (SimulASR)\cite{fiscus2006multiple,guoqiang2024simul} and Simultaneous Machine Translation (SimulMT)\cite{ma2018stacl}. These auxiliary tasks provide additional contextual information, which helps improve the performance of the primary task.
While MTL has demonstrated advantages in enhancing overall model effectiveness, it also introduces substantial challenges, particularly the potential for the conflicts between task optimizations.

Within the MTL framework, auxiliary tasks are not merely supportive, they can also harm the performance of the primary task. The differing optimization objectives among tasks result in divergent gradient directions during back-propagation. As training processes, these discrepancies in gradient directions tend to intensify. Researchers \cite{yu2020gradient} define \textit{Gradient Conflict} as the scenario where the angle between gradient directions of different tasks exceeds 90$^{\circ}$. Such conflicts can negatively impact the optimization process. The PCGrad strategy \cite{yu2020gradient} is proposed to address this issue, which mitigates gradient conflicts by projecting the conflicting gradients onto the normal vectors of the primary. Although PCGrad is a simple and effective method that tries to alleviate conflicts, simply introducing PCGrad in SimulST brings worse results and reveals two major drawbacks, i.e., conflict masking and inefficient memory usage \cite{smith2017don}.

\definecolor{qgreen}{RGB}{226,240,217}
\definecolor{qyellow}{RGB}{255,242,204}
\definecolor{qorange}{RGB}{248,203,173}
\definecolor{qblue}{RGB}{189,215,238}
\definecolor{qmidgreen}{RGB}{169,209,142}
\definecolor{qmidyellow}{RGB}{255,217,102}
\definecolor{qpenorange}{RGB}{238,140,69}
\definecolor{qpenblue}{RGB}{68,114,196}
\definecolor{qkgreen}{RGB}{118,176,79}
\definecolor{qkorange}{RGB}{255,192,0}

\begin{figure}

\begin{tikzpicture}
\tikzstyle{texttitle} = [font=\scriptsize,align=center]
\tikzstyle{textonly} = [font=\scriptsize,align=left]
\tikzstyle{sublayer} = [rectangle,draw,minimum width=2.4cm,line width=0.8pt,rounded corners=5pt,align=center,inner sep=3pt,minimum height=0.8cm,font=\scriptsize];
\tikzstyle{subsublayer} = [rectangle,draw,minimum width=1.8cm,rounded corners=3pt,align=center,inner sep=3pt,minimum height=0.8cm,font=\scriptsize];
\tikzstyle{mycircle} = [circle,draw,minimum width=0.4cm,rounded corners=3pt,align=center,inner sep=3pt,minimum height=0.3cm,font=\small];
\tikzstyle{background} = [rectangle,draw,rounded corners=5pt,minimum width=2.4cm,minimum height=2.1cm,font=\footnotesize,align=center];
\tikzstyle{background_legend} = [rectangle,draw,rounded corners=3pt,minimum width=6cm,minimum height=0.5cm,font=\footnotesize,align=center];
\tikzstyle{back} = [rectangle,rounded corners=5pt,minimum width=3.2cm,minimum height=0.8cm,font=\footnotesize,align=center];
\tikzstyle{back1} = [rectangle,rounded corners=5pt,minimum width=8.4cm,minimum height=1.05cm,font=\footnotesize,align=center];
\tikzstyle{back2} = [rectangle,rounded corners=5pt,minimum width=8.4cm,minimum height=2.3cm,font=\footnotesize,align=center];
\node[textonly](GT) at (-4.2,0) {};
\node[back1,draw=black](GG) at (0,0){};
\node[sublayer,color=qmidgreen,fill=qgreen,fill opacity=1](SimulST) at ([xshift=-1.4cm, yshift=0.525cm]GG.south) {};
\node[texttitle, color=black] at ([yshift=-0.28cm]SimulST.north) {SimulST};
\node[texttitle] at ([yshift=0.24cm,xshift=0.05cm]SimulST.south) {
    \textcolor{black}{(}
    \textcolor{qpenorange}{\textbf{0.5}}
    \textcolor{black}{,}
    \textcolor{qpenorange}{\textbf{0.4}}
    \textcolor{black}{,}
    \textcolor{qpenblue}{\textbf{0.7}}
    \textcolor{black}{,}
    \textcolor{qpenblue}{\textbf{0.4}}
    \textcolor{black}{)}
  };
\node[sublayer,color=qmidyellow,fill=qyellow,fill opacity=1](SimulASR) at ([xshift=2.9cm, yshift=0.525cm]GG.south) {};
\node[texttitle, color=black] at ([yshift=-0.28cm]SimulASR.north) {SimulASR};
\node[texttitle] at ([yshift=0.24cm,xshift=0.05cm]SimulASR.south) {
    \textcolor{black}{(}
    \textcolor{qpenorange}{\textbf{0.9}}
    \textcolor{black}{,}
    \textcolor{qpenorange}{\textbf{0.8}}
    \textcolor{black}{,}
    \textcolor{qpenblue}{\textbf{-0.9}}
    \textcolor{black}{,}
    \textcolor{qpenblue}{\textbf{0.7}}
    \textcolor{black}{)}
  };
\node[texttitle, color=black] (J1) at ([xshift=0.75cm,yshift=0.725cm]GG.south) {\textbf{0.42}};
\draw[stealth-stealth,color=black,line width=0.5pt] ([xshift=-0.8cm, yshift=-0cm] J1.south) -- ([xshift=0.8cm, yshift=-0cm] J1.south);
\node[font=\footnotesize] at ([yshift=-0.18cm]J1.south) {\usym{1F5F8}};

\node[texttitle, color=black] at ([yshift=0.3cm]SimulST.north) {Primary Task};
\node[texttitle, color=black] at ([yshift=0.3cm]J1.north) {Cos Similarity};
\node[texttitle, color=black] at ([yshift=0.3cm]SimulASR.north) {Auxiliary Task};
\node[texttitle, color=black] at ([xshift=-0.8cm]SimulST.west) {Coarse-grained\\Gradients};
\node[back2,draw=black  ](LG) at ([yshift=-1.3cm]GG.south){};
\node[background,densely dashed,color=qkgreen,line width=0.7pt](ST) at ([xshift=-1.4cm, yshift=-1.15cm]LG.north) {};
\node[subsublayer,color=qpenorange,fill=qorange,fill opacity=1](ST_Encoder) at ([yshift=-0.5cm]ST.north) {};
\node[texttitle, color=black] at ([yshift=-0.28cm]ST_Encoder.north) {Encoder};
\node[texttitle] at ([yshift=0.24cm]ST_Encoder.south) {
    \textcolor{black}{(}
    \textcolor{qpenorange}{\textbf{0.5}}
    \textcolor{black}{,}
    \textcolor{qpenorange}{ \textbf{0.4}}
    \textcolor{black}{)}
  };
\node[subsublayer,color=qpenblue,fill=qblue,fill opacity=1](ST_Decoder) at ([yshift=0.5cm]ST.south) {};
\node[texttitle, color=black] at ([yshift=-0.28cm]ST_Decoder.north) {Decoder};
\node[texttitle] at ([yshift=0.24cm]ST_Decoder.south) {
    \textcolor{black}{(}
    \textcolor{qpenblue}{\textbf{0.7}}
    \textcolor{black}{,}
    \textcolor{qpenblue}{ \textbf{0.4}}
    \textcolor{black}{)}
  };
\node[background,densely dashed,color=qkorange,line width=0.7pt](ASR) at ([xshift=2.9cm, yshift=-1.15cm]LG.north) {};
\node[subsublayer,color=qpenorange,fill=qorange,fill opacity=1](ASR_Encoder) at ([yshift=-0.5cm]ASR.north) {};
\node[texttitle, color=black] at ([yshift=-0.28cm]ASR_Encoder.north) {Encoder};
\node[texttitle] at ([yshift=0.24cm]ASR_Encoder.south) {
    \textcolor{black}{(}
    \textcolor{qpenorange}{\textbf{0.9}}
    \textcolor{black}{,}
    \textcolor{qpenorange}{\textbf{0.8}}
    \textcolor{black}{)}
  };
\node[subsublayer,color=qpenblue,fill=qblue,fill opacity=1](ASR_Decoder) at ([yshift=0.5cm]ASR.south) {};
\node[texttitle, color=black] at ([yshift=-0.28cm]ASR_Decoder.north) {Decoder};
\node[texttitle] at ([yshift=0.24cm]ASR_Decoder.south) {
    \textcolor{black}{(}
    \textcolor{qpenblue}{\textbf{-0.9}}
    \textcolor{black}{,}
    \textcolor{qpenblue}{\textbf{0.7}}
    \textcolor{black}{)}
  };
\node[texttitle, color=black] (J2) at ([xshift=0.75cm,yshift=-0.42cm]LG.north) {\textbf{0.77}};
\node[font=\footnotesize] at ([yshift=-0.18cm]J2.south) {\usym{1F5F8}};
\draw[stealth-stealth,color=black,line width=0.5pt] ([xshift=-0.8cm, yshift=-0cm] J2.south) -- ([xshift=0.8cm, yshift=0cm] J2.south);

\node[texttitle, color=red](J3) at ([xshift=0.75cm,yshift=0.9cm]LG.south){\textbf{-0.35}};
\draw[stealth-stealth,color=red,line width=0.5pt] ([xshift=-0.8cm, yshift=0cm] J3.south) -- ([xshift=0.8cm, yshift=0cm] J3.south);
\node[font=\scriptsize,color=red] at ([yshift=-0.18cm]J3.south) {\usym{1F5F4}};
\node[texttitle, color=red] at ([yshift=-0.4cm]J3.south) {Gradient Conflict};
\node[texttitle, color=black] at ([xshift=-0.8cm]ST.west) {Fine-grained\\Gradients};
\draw[->,color=qmidgreen,line width=2pt] ([yshift=-0.05cm] SimulST.south) -- ([yshift=0.1cm] ST_Encoder.north);
\draw[->,color=qmidyellow,line width=2pt] ([yshift=-0.05cm] SimulASR.south) -- ([yshift=0.1cm] ASR_Encoder.north);
\draw[-stealth,color=black,line width=0.7pt] (ST_Encoder.south) -- (ST_Decoder.north);
\draw[-stealth,color=black,line width=0.7pt] (ASR_Encoder.south) -- (ASR_Decoder.north);

\end{tikzpicture}
\caption{Comparison gradient conflicts of different model granularity using cosine similarity. Coarse-grained model-level gradients show no conflicts (top). However, the finer-grained level reveals conflicts between the decoders (bottom), illustrating the concept of gradient conflict masking.}
\label{fig1}
\end{figure}
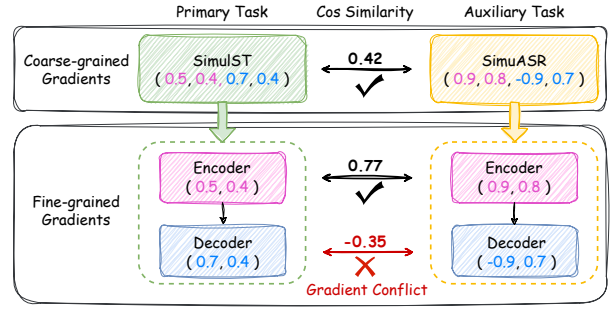

 Fig. \ref{fig1} shows a conflict masking situation. Although there seems to be no conflict in the model (coarse-grained situation), the conflict actually exists when we observe the model at a finer-grained granularity. This masking phenomenon means that the conflict issue is not really resolved. Additionally, introducing similarity calculations incurs significant computational costs, especially when concatenating the gradients of all parameters into a single, extremely long vector, occupying GPU memory during vector storing and processing.

Both of these issues arise from trying to resolve conflicts at a coarse-grained model level. By partitioning the model into finer-grained modules, we can more accurately detect and address conflict locations, significantly reducing unnecessary resource consumption.
We introduce \textit{\textbf{M}odular \textbf{G}radient \textbf{C}onflict \textbf{M}itigation (\textbf{MGCM})} strategy which specifically targets conflict resolution at the module level. 
MGCM effectively mitigates optimization conflicts, reduces GPU memory requirements, and improves both training efficiency and overall model performance.
Our contributions are as follows:
\begin{itemize} 
\item We address the challenge of gradient conflicts being masked by numerical inaccuracies, ensuring more precise optimization. 
\item MGCM introduces only a negligible memory overhead, with an increase of only 0.2 GB for a model with 200M parameters. 
\item Our proposed systems\footnote{The code is available at \href{https://github.com/jianguo0319/MGCM}{https://github.com/jianguo0319/MGCM.}} consistently outperform baseline strategies in both offline and simultaneous settings, demonstrating superior accuracy and efficiency. \end{itemize}

\section{Method}
In this section, we provide a detailed explanation of the method. We begin by analyzing the limitations of resolving gradient conflicts at the model level, followed by the introduction of our proposed modular gradient conflict mitigation strategy.



\subsection{Limitations of Model-level Gradient Conflict Resolution}

Projecting Conflicting Gradients (PCGrad) \cite{yu2020gradient} is an optimization method designed to handle gradient conflicts in MTL and has been applied in various fields. To ensure all parameters are optimized together during back-propagation, PCGrad concatenates the gradients of all into a long model-level vector $G$. Given the total number of the model parameters $n$, $g^i$ denotes the gradient of the $i$-th parameter, $G$ can be described as:
\begin{align}
{G} = g^1 \oplus g^2 \oplus \cdots \oplus g^n
\label{con}
\end{align}

Then it calculates the cosine similarity between the primary and the auxiliary task gradient and reduces conflicts by projecting conflicting gradients onto a plane orthogonal to the primary. When applied to the SimulST task within the MTL framework, PCGrad reveals two drawbacks, both stem from the way it concatenates gradients for all model parameters.
\subsubsection{Gradient Conflict Masking}

We monitor a contradicting situation in that PCGrad often computes near-zero cosine similarity, suggesting that task optimization directions are nearly orthogonal \cite{chen2024pareto, anastasopoulos2018tied,wang2020bridging}. Further analysis shows that gradient conflicts are usually confined to specific model components like attention mechanisms, and become evident only after further optimization. Unfortunately, these conflicts are usually masked because when calculating cosine similarity via such a large model-level gradient vector, more contributions from the non-conflicting modules are dominating.

\subsubsection{GPU Memory Consumption} 
Since PCGrad utilizes the model-level gradient vector, the length of this vector increases as the number of model parameters grows. Calculating similarity for such a large vector requires approximately as 2.5$\times$ as the input vector's GPU memory, undoubtedly leading to significant memory overhead. Additionally, the large vector size makes it susceptible to overflow when calculating in lower precision, forcing a switch to higher precision, and further exacerbating memory consumption.

These issues motivate us to explore a more refined and modular approach that operates at a finer-grained level, aiming to accurately detect and resolve conflicts where they actually occur, meantime improving memory efficiency and maintaining the effectiveness of conflict mitigation.

\subsection{MGCM: Modular Gradient Conflict Mitigation}

To address the limitations of PCGrad, we propose the MGCM strategy as outlined in Algorithm \ref{algor}. MGCM focuses on detecting and mitigating conflicts at the module level, thereby overcoming issues associated with model-level gradient approaches.

\definecolor{qorange}{RGB}{237,125,49}
\definecolor{qgreen}{RGB}{112,173,71}
\definecolor{qblue}{RGB}{68,114,196}

\begin{figure}[t]
    \centering
    \subfigure[Detection]{
    \label{fig2a}
        \begin{tikzpicture}[scale=1]

            \draw[qorange, thick,line width=1pt] (0.3,0) arc[start angle=0, end angle=45, radius=0.3];
            \draw[->, thick,line width=1.5pt, qgreen] (0.017,0) -- (1.1,1.2);
            \draw[->, thick,line width=1.5pt, qblue] (0,0) -- (1.5,0);

            \draw[qorange, thick,line width=1pt] (2.7,0) arc[start angle=0, end angle=120, radius=0.2];
            \draw[->, thick,line width=1.5pt, qgreen] (2.52,0) -- (1.7,1.2);
            \draw[->, thick,line width=1.5pt, qblue] (2.5,0) -- (4,0);

        \end{tikzpicture}
    }
    \hspace{0.3cm}
    \subfigure[Projection]{
    \label{fig2b}
        \begin{tikzpicture}[scale=1]
            \draw[->, dashed,line width=1.5pt, thick, red] (-1.0,1.2) -- (-0.04,1.2);
            \draw[qorange, thick,line width=1pt] (0.25,0) arc[start angle=0, end angle=90, radius=0.25];
            \draw[->, thick,line width=1.5pt, qgreen] (0,0) -- (-1.2,1.2);
            \draw[->, dashed, thick,line width=1.5pt, qgreen] (0,0) -- (0,1.2);
            \draw[->, thick,line width=1.5pt, qblue] (-0.02,0) -- (1.5,0);
            \node at (-0.50,1.45) {\footnotesize{Projection}};
        \end{tikzpicture}
    }
    \caption{Gradient detection and projection in MTL. (a) detects conflicts by calculating the cosine similarity between the primary and auxiliary tasks, illustrating whether the state is without or with conflicts. (b) captures the conflicts and demonstrates the projection process to mitigate them.}
    \label{fig2}
\end{figure}
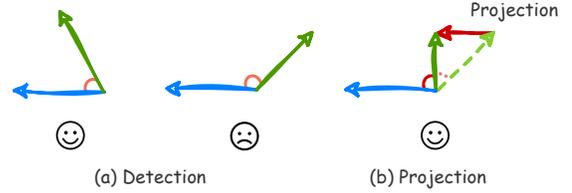

\subsubsection{Modularization} 

To precisely identify where gradient conflicts occur, we first divide the model into fine-grained modules. The challenge lies in determining how fine the level of granularity should be. The key requirement is that after modularization, the model can still correctly execute forward and backward propagation without disrupting its overall optimization. Thus, a \textit{Module} is defined as the minimal unit that can independently compute gradients during optimization, ensuring that modularization identifies conflicts at a manageable level without compromising the model's functionality or performance.
Specifically, a Transformer-based model can be viewed as consisting of the 3 key components, each of which can be divided into a few modules $M$:
\begin{itemize}
    \item \textbf{Layer Norm} comprises the $LN$ module, i.e. the scaling and shifting parameters within the normalization layer.
    \item \textbf{Feed Forward Network} includes weights for the two fully connected layers, represented by $W_{1}$ and $W_{2}$ modules.
    \item \textbf{Attention Mechanism} consists of weights for the query ($Q$), key ($K$), and value ($V$) of the attention head and the linear output weight ($O$).
\end{itemize}

Modularization function $Modularize(\cdot)$ extracts sub-gradients for each module. It can be defined as Eq. \eqref{module}:
\begin{align}
Modularize({G}) = [\mathbf{g}^{1},\mathbf{g}^{2}, \cdots, \mathbf{g}^{m}]
\label{module}
\end{align}
where $m$ denotes the total number of modules and $\mathbf{g}^{j}$ represents the gradient associated with the $j$-th module. In the subsequent equations, all instances of $\textbf{g}$ refer to the gradients of the tasks within the same module.
\subsubsection{Conflict Detection and Projection} 

MGCM detects conflicts by calculating the cosine similarity between gradients within a module. Given that $\mathbf{g}_p$ denotes the primary task gradient and $\mathbf{g}_a$ denotes the auxiliary, conflict detection function ${CosSim}(\cdot, \cdot)$ is shown as:
\begin{align}
{CosSim}(\mathbf{g}_p, \mathbf{g}_a) = \frac{\mathbf{g}_p \cdot \mathbf{g}_a }{||\mathbf{g}_p|| \  ||\mathbf{g}_a||}
\label{cosine}
\end{align}

Fig. \ref{fig2a} illustrates the detection process. It shows ideal gradient directions within a module (left), where auxiliary tasks aid in convergence and enhance the model’s generalization by expanding the search space \cite{wang2020bridging,tang2022unified,crammer2012learning}.
When conflict arises (right), MGCM will employ a projection approach as shown in Fig. \ref{fig2b}. The auxiliary gradients will be projected onto the plane orthogonal to the primary for aligning, described by Eq. \eqref{pc}:
\begin{align}
    \mathbf{g}_a = \mathbf{g}_a - \frac{\mathbf{g}_p \cdot \mathbf{g}_a}{||\mathbf{g}_p||^2} \cdot \mathbf{g}_p
    \label{pc}
\end{align}

The model parameters are then updated utilizing these adjusted gradients, minimizing conflicts at the module level. By focusing on module-level conflicts, MGCM effectively addresses gradient conflict masking observed in PCGrad. Meanwhile, the modular approach leads to an incidental benefit of GPU memory saving, as it avoids the need to compute long vectors. The resulting memory savings are quantified and discussed in Section \ref{gpu_consum}.

\begin{algorithm}[t]
\SetKwFunction{Union}{Union}\SetKwFunction{Modularize}{Modularize}
\small
\DontPrintSemicolon

  \KwInit{0: SimulST, 1: SimulASR, 2: SimulMT}
   \KwInput{gradients of tasks ${G}_{0}, {G}_{1}, {G}_{2}$ and modules $M$ }
   \KwOutput{total gradients of the model $G$}
   initialization: {${G} = [\ ]$}; \\
  \tcc*[r]{Modularization}
  ${G}_{i}\leftarrow$ \Modularize{${G}_{i}$}\ $ \; \forall \ i \in \{0,1,2\}$;\\
  \tcc*[r]{Conflict Detection and Projection}
  \SetAlgoLined\SetArgSty{}
  \ForAll{elements of $M$}{
    ${\mathbf{g}}_{i}\leftarrow{G}_{i} $ [element] \ $ \forall \ i \in \{0,1,2\}$;\\
    
    \If{${\mathbf{g}}_{0} \cdot {\mathbf{g}}_{1} < 0$}{
        ${\mathbf{g}}_{1}\leftarrow{\mathbf{g}}_{1} - \displaystyle\frac{{\mathbf{g}}_{1} \cdot {\mathbf{g}}_{0}}{||{\mathbf{g}}_{0}||^2} \cdot {\mathbf{g}}_{0}$; \tcc*[r]{SimulASR} 
    }

    \If{${\mathbf{g}}_{0} \cdot {\mathbf{g}}_{2} < 0$}{
        ${\mathbf{g}}_{2}\leftarrow{\mathbf{g}}_{2} - \displaystyle\frac{{\mathbf{g}}_{2} \cdot {\mathbf{g}}_{0}}{||{\mathbf{g}}_{0}||^2} \cdot {\mathbf{g}}_{0}$;\tcc*[r]{SimulMT}
    }
    
    ${G}.append ({\mathbf{g}}_{0}+{\mathbf{g}}_{1}+{\mathbf{g}}_{2})$;
  }
  \textbf{return}\  $G$;
\caption{Modular Gradient Conflict Mitigation}
\label{algor}
\end{algorithm}

\section{Experiments and Analysis}
\label{sec:experiments}

\subsection{Datasets and Pre-processing}
\label{ssec:datasets}

We construct experiments on the MuST-C dataset \cite{di2019must}. MuST-C is a multilingual ST corpus extracted from the TED talks. We test our method on the English (En) to German (De) portion of the MuST-C v1 corpus\footnote{\href{https://ict.fbk.eu/must-c}{https://ict.fbk.eu/must-c}}, which contains about 400 hours of speech and 230K utterances. We select the model based on the dev set (1,408 utterances) and report results on the tst-COMMON set (2,641 utterances).
For pre-processing, we use original WAV files sampled at 16 kHz. The text data is processed using SentencePiece\footnote{\href{https://github.com/google/sentencepiece}{https://github.com/google/sentencepiece}} \cite{kudo2018sentencepiece} to generate a unigram vocabulary of size 10,000. Due to the repetitive nature of roots and affixes in both English and German, we employ a unified dictionary.

\subsection{Model Settings}
\label{ssec:setting}

We use SimulST as the primary task, SimulASR and SimulMT as auxiliary tasks in the MTL framework. The models are conducted using the fairseq toolkit \cite{wang2020fairseq}. To minimize the difficulty of speech feature extraction, we use the pre-trained wav2vec 2.0 \cite{baevski2020wav2vec} small model for feature extraction. 
For modeling, we utilize a Transformer-based structure, which consists of 6 encoder layers and 6 decoder layers. Each layer contains 8 attention heads, with a hidden layer size of 768 units and an FFN layer size of 2048 units. For simultaneous generating, we employ a differentiable policy and details can refer to Zhang and Feng\cite{zhang2023end}.
Additionally, SimulEval toolkit\footnote{\href{https://github.com/facebookresearch/SimulEval}{https://github.com/facebookresearch/SimulEval}} \cite{ma2020simuleval} is used to evaluate both latency and model performance. The latency is evaluated using Average Lagging (AL) metric\cite{ma2018stacl}, while the performance is measured by SacreBLEU metric\footnote{\href{https://github.com/mjpost/sacrebleu}https://github.com/mjpost/sacrebleu} \cite{post2018call}.

\definecolor{color1}{RGB}{255,000,000}
\definecolor{color2}{RGB}{046,117,182}
\definecolor{color3}{RGB}{112,173,071}
\definecolor{color4}{RGB}{237,125,049}
\definecolor{color5}{RGB}{158,072,014}
\definecolor{color6}{RGB}{165,165,165}
\definecolor{color7}{RGB}{255,192,000}
\definecolor{color8}{RGB}{112,048,160}
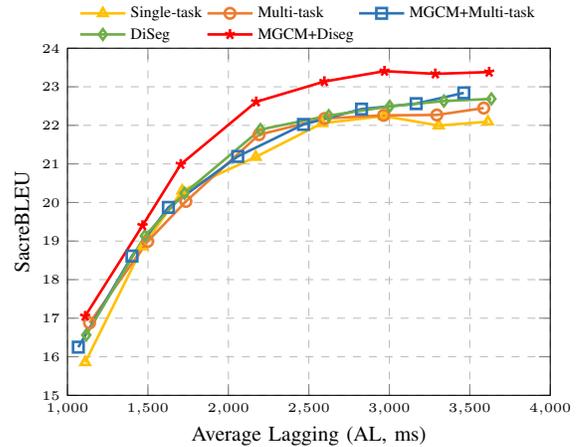
\begin{figure}[t]
\centering
  \begin{tikzpicture}
    \footnotesize{
      \begin{axis}[
        fill opacity=1,
        fill=orange,
        ymajorgrids,
        xmajorgrids,
        grid style=dashed,
        width=8cm,
        height=6.5cm,
        legend columns=3,
        legend entries={
         Single-task, Multi-task, MGCM+Multi-task, DiSeg, MGCM+Diseg
        },
        legend style={fill opacity=0,text opacity =1,
          draw=none,
          line width=1pt,
          legend cell align=left,
        },
        legend style={
        at={(0.5, 1.15)}, anchor=north,
        nodes={scale=0.75, transform shape}
        },
        xmin=1000, xmax=4000,
        ymin=15, ymax=24,
        xtick={1000,1500,...,4000},
        ytick={15,16,...,24},
        xticklabel style={font=\tiny},
        yticklabel style={font=\tiny},
        xlabel=\footnotesize{Average Lagging (AL, ms)},
        ylabel=\footnotesize{SacreBLEU},
        ylabel style={yshift=-0.1em},
        xlabel style={yshift=0.1em},
        scaled ticks=false,
        ]
        
    \addplot[color7, mark=triangle, line width=1pt] 
    file {data/data_table1/singletask.txt}; 
    \addplot[color4, mark=o, line width=1pt] 
    file {data/data_table1/multitask.txt}; 
    \addplot[color2, mark=square, line width=1pt] 
    file {data/data_table1/mgcm.txt}; 
    \addplot[color3, mark=diamond, line width=1pt] 
    file {data/data_table2/diseg.txt}; 
    \addplot[color1, mark=star, line width=1pt] 
    file {data/data_table3/mgcm.txt}; 
    \end{axis}
    }
  \end{tikzpicture}
\caption{We initially test a Transformer-based model (with a differentiable simultaneous strategy) across three scenarios: single-task, multi-task, and MGCM-enhanced multi-task learning. Subsequently, we employ the more robust Diseg model as a baseline to further evaluate the MGCM method.}
\label{fig3}
\end{figure}

\subsection{Main Results}
\label{sec:results}

As depicted in Fig. \ref{fig3}, it can be observed that a simple MTL method leads to performance improvements in SimulST across all latency levels. However, despite the significant resource consumption, the improvement is relatively modest, with the BLEU scores increasing by only 0.4 points at high latencies. With the introduction of MGCM, the improvements remain marginal at low latencies but become more pronounced at medium and high latencies, with BLEU scores increasing by 1 point.

To further assess the generalization capability of the MGCM method, we use the more powerful open-source DiSeg model\footnote{Despite extensive efforts to replicate, we do not achieve the performance reported in the original paper. The results here reflect our best attempt.} \cite{zhang2023end} as the baseline. Notably, the MGCM mechanism substantially enhances overall model performance under medium and high latency conditions. These improvements are particularly pronounced, indicating that MGCM effectively leverages the extended time span to integrate better and utilize contextual information.

We observe a general trend of suboptimal performance across most models at low latency. This decline in performance can be attributed to the substantial information loss that occurs when limited processing time restricts the ability to capture and integrate necessary contextual cues. However, at medium to high latency levels, the MGCM mechanism not only mitigates these limitations but also significantly strengthens the generalization capabilities of auxiliary tasks. Consequently, this enhancement leads to substantial improvements in the performance of SimulST tasks, thereby underscoring the effectiveness of the MGCM approach in scenarios that allow for more extensive processing time.

\subsection{Ablation Study}

We then validate the effectiveness of the proposed MGCM method using the DiSeg structure as a unified baseline. We compare MGCM against other gradient conflict mitigation techniques, including the model-level method PCGrad and a simple approach we term ``Discard,'' where conflicting modules are simply discarded without any further operation.

While the Discard method performs similarly to MGCM under low-latency conditions as illustrated in Fig. \ref{fig4}, it falls short in medium to high-latency scenarios. The discarded gradients do not contribute to the model’s generalization, resulting in performance that is nearly equivalent to the baseline, which confirms the necessity of conflict addressing.

PCGrad even underperforms compared to the baseline, as shown in Fig. \ref{fig4}. This is because PCGrad addresses conflicts on a model scale, and by the time a conflict is detected, local conflicts may have already escalated. Additionally, projecting the model-level gradient negatively impacts the optimization of non-conflicting modules, leading to further performance degradation.

In contrast, MGCM matches baseline performance at low latency but exhibits significant improvements at medium and high latencies, demonstrating its superior capability to handle more complete information. This advantage is even more pronounced in the offline setting, as shown in Table \ref{tab1}, where we observe substantial gains under both Greedy (+0.68 BLEU) and Beam5 (+0.63 BLEU) decoding strategies.

\definecolor{color1}{RGB}{255,000,000}
\definecolor{color2}{RGB}{046,117,182}
\definecolor{color3}{RGB}{112,173,071}
\definecolor{color4}{RGB}{237,125,049}
\definecolor{color5}{RGB}{158,072,014}
\definecolor{color6}{RGB}{165,165,165}
\definecolor{color7}{RGB}{255,192,000}
\definecolor{color8}{RGB}{112,048,160}
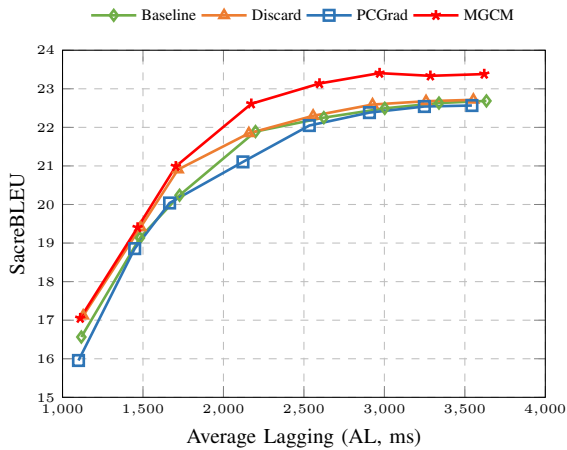
\begin{figure}[t]
\centering
  \begin{tikzpicture}
    \footnotesize{
      \begin{axis}[
        fill opacity=1,
        fill=orange,
        ymajorgrids,
        xmajorgrids,
        grid style=dashed,
        width=8cm,
        height=6.5cm,
        legend columns=-1,
        legend entries={
            Baseline, Discard, PCGrad, MGCM
        },
        legend style={fill opacity=0,text opacity =1,
          draw=none,
          line width=1pt,
          legend cell align=left,
        },
        legend style={
        at={(0.5, 1.15)}, anchor=north,
        nodes={scale=0.75, transform shape}
        },
        xmin=1000, xmax=4000,
        ymin=15, ymax=24,
        xtick={1000,1500,...,4000},
        ytick={15,16,...,24},
        xticklabel style={font=\tiny},
        yticklabel style={font=\tiny},
        xlabel=\footnotesize{Average Lagging (AL, ms)},
        ylabel=\footnotesize{SacreBLEU},
        ylabel style={yshift=-0.1em},
        xlabel style={yshift=0.1em},
        scaled ticks=false,
        ]
    \addplot[color3,mark=diamond, line width=1pt] 
    file {data/data_table2/diseg.txt}; 
    \addplot[color4,mark=triangle,line width=1pt]
    file {data/data_table2/discard.txt};
    \addplot[color2,mark=square, line width=1pt] 
    file {data/data_table2/pcgrad.txt};
    \addplot[color1, mark=star, line width=1pt] 
    file {data/data_table2/mgcm.txt};  
    \end{axis}
    }
  \end{tikzpicture}
\caption{The performance of various gradient conflict mitigation methods, with all models utilizing the Diseg model as the baseline.}
\label{fig4}
\end{figure}

\begin{table}[t]
\centering
\small
\caption{Performance of each model in offline translation tasks.}
\begin{tabular}{L{1.5cm}C{1.5cm}C{1.5cm}}\toprule
\multirow{2}{*}{\textbf{Methods}}      & \multicolumn{2}{c}{\textbf{BLEU}} \\\cmidrule(lr){2-3}
        & Greedy         & Beam5           \\\midrule
DiSeg   & 22.95          & 24.51           \\
PCGrad  & 22.75          & 24.28           \\
Discard & 22.84          & 24.43           \\
MGCM    & \textbf{23.63} & \textbf{25.14}  \\\bottomrule
\end{tabular}

\label{tab1}
\end{table}

\subsection{Conflict Probability Analysis}
We explore the probability of conflicts between primary and auxiliary tasks to identify where conflicts actually occur. The results reveal a stable conflict probability of around 50\% in the attention (Attn) mechanism for both SimulMT and SimulST, while conflicts are minimal in the FFN and LN components. Therefore, Fig. \ref{fig5} displays only the conflict probabilities between SimulASR and SimulST.

As illustrated in Fig. \ref{fig5}, the Attn mechanism shows high conflict probability across all layers, while FFN and LN exhibit higher probabilities only in specific layers. The high conflict probability in the Attn mechanism is due to different attention heads focusing on varying parts of the input sequence, leading to frequent disagreements in gradient updates.

Additionally, the conflict probabilities for FFN and LN increase sharply in Layer 6. We hypothesize that this increase is due to Layer 6 transitioning from encoding general features to preparing task-specific representations. This shift amplifies the divergence between the task-specific requirements of FFN and LN components, resulting in higher conflict probabilities as the model aligns with decoder inputs.

\definecolor{color1}{RGB}{075,102,173}
\definecolor{color2}{RGB}{098,190,166}
\definecolor{color3}{RGB}{253,186,107}
\definecolor{color4}{RGB}{235,096,070}
\begin{figure}[t]
\centering
  \begin{tikzpicture}
  \tikzstyle{textonly} = [font=\scriptsize,align=left]
  \node[textonly] at (5,0) {};
    \footnotesize{
      \begin{axis}[
        fill opacity=1,
        fill=orange,
        ymajorgrids,
        xmajorgrids,
        grid style=dashed,
        width=0.45\textwidth,
        height=0.3\textwidth,
        legend columns=-1,
        legend entries={
         Attn, FFN, LN
        },
        legend style={fill opacity=0.5,text opacity =1,
          draw=none,
          line width=1pt,
        },
        legend style={
        at={(0.5,1.15)}, anchor=north,
        nodes={scale=0.75, transform shape}
        },
        xmin=1, xmax=12,
        ymin=0, ymax=0.6,
        xtick={1,2,3,...,12},
        ytick={0,0.1,...,0.6},
        xticklabel style={font=\tiny},
        yticklabel style={font=\tiny},
        xlabel=\footnotesize{Layer},
        ylabel=\footnotesize{Conflict Probability},
        ylabel style={yshift=-0.1em},
        xlabel style={yshift=0.1em},
        scaled ticks=false,
        ]

    \addplot[color2,mark=*, line width=1pt] 
    file {data/data_conflict/asr_attn.txt}; 
    \addplot[color3, mark=square, line width=1pt] 
    file {data/data_conflict/asr_ffn.txt}; 
    \addplot[color4, mark=diamond, line width=1pt] 
    file {data/data_conflict/asr_ln.txt}; 
    \end{axis}
    }
    \end{tikzpicture}
\caption{The probability of conflicts between SimulST and SimulASR at the component level (Attn, FFN, LN) across different Transformer layers. Layers 1-6 represent the encoders, and layers 7-12 represent the decoders.}
\label{fig5}
\end{figure}
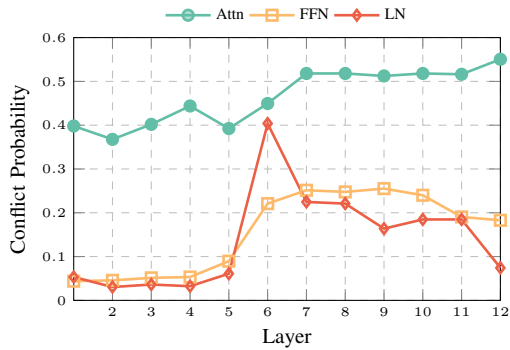

\subsection{Statistical Significance Test} 
 We further ensure the reliability of MGCM by employing statistical significance tests. When the p-value is greater than 0.05, it indicates that the observed differences between the methods are probably due to randomness. We perform tests between online and offline inference settings and find that the maximum p-value is 0.034 in En-De. All values are well below the 0.05 threshold, ensuring that the performance improvements achieved by MGCM are statistically significant and not due to random fluctuations.

\subsection{GPU Memory Consumption}
\label{gpu_consum}

As shown in Table \ref{tab2}, we quantified the GPU memory savings achieved by MGCM. PCGrad processes the full model gradients, whereas MGCM handles only a fraction, no more than 1/20 of the total gradient at a time. In similarity computation, PCGrad requires 4.5$\times$ the model's total GPU memory, while MGCM only consumes 0.225$\times$. Therefore, MGCM saves approximately 4.275$\times$ the GPU memory compared to PCGrad.

At FP32 precision, MGCM saves roughly 18MB of GPU memory every 1M parameters. For a model with 0.2B parameters, MGCM can save around 3.57GB of memory compared with PCGrad. As the number of model parameters increases, the memory savings scale accordingly, making MGCM more efficient for larger models.



\begin{table}[t]
\centering
\small
\caption{Comparison of the extra memory usage for MGCM and PCGrad at different Parameters.}
\begin{tabular}{L{1.3cm}C{1.3cm}C{1.3cm}C{2.1cm}}\toprule
Params (Billion) & PCGrad (GB) & MGCM (GB) & Memory Saving (Percent) \\\midrule
0.2 B  & 3.77         & 0.20        & 94.69\%          \\
0.5 B  & 9.41         & 0.48        & 94.90\%           \\ 
1.0 B  & 18.81        & 0.95        & 94.95\%            \\\bottomrule
\end{tabular}
\label{tab2}
\end{table}

\section{Conclusion}
\label{sec:conclusion}

We address the limitations of model-level gradient conflict resolution in SimulST tasks within MTL by introducing MGCM. Through extensive experiments, we demonstrate that MGCM effectively mitigates gradient conflicts at the modular level, offering significant advantages over existing methods like PCGrad. MGCM not only resolves the issue of gradient conflict masking observed in model-level approaches but also achieves substantial GPU memory savings, particularly as model sizes continue to grow. Our comparative analysis shows that MGCM outperforms another method across various latency settings in SimulST tasks, especially under medium to high latency conditions where more complete information is available. MGCM’s potential to better utilize extended processing time makes it a robust solution for tasks requiring real-time translation.

\bibliographystyle{IEEEtran}
\bibliography{refs}

\end{document}